# Compact Hyperplane Hashing with Bilinear Functions

**Wei Liu**[†] **Jun Wang**[‡] **Yadong Mu**[†] **Sanjiv Kumar**[§] **Shih-Fu Chang**[†]

[†]Columbia University, New York, NY 10027, USA {wliu,muyadong,sfchang}@ee.columbia.edu

[‡]IBM T. J. Watson Research Center, Yorktown Heights, NY 10598, USA wangjun@us.ibm.com

[§]Google Research, New York, NY 10011, USA sanjivk@google.com

## Abstract

Hyperplane hashing aims at rapidly searching nearest points to a hyperplane, and has shown practical impact in scaling up active learning with SVMs. Unfortunately, the existing randomized methods need long hash codes to achieve reasonable search accuracy and thus suffer from reduced search speed and large memory overhead. To this end, this paper proposes a novel hyperplane hashing technique which yields compact hash codes. The key idea is the bilinear form of the proposed hash functions, which leads to higher collision probability than the existing hyperplane hash functions when using random projections. To further increase the performance, we propose a learning based framework in which the bilinear functions are directly learned from the data. This results in short yet discriminative codes, and also boosts the search performance over the random projection based solutions. Large-scale active learning experiments carried out on two datasets with up to one million samples demonstrate the overall superiority of the proposed approach.

## 1. Introduction

Fast approximate nearest neighbor search arises commonly in a variety of domains and applications due to massive growth in data that one is confronted with. An attractive solution to overcome the speed bottleneck that an exhaustive linear scan incurs is the use of algorithms from the *Locality-Sensitive Hashing* (LSH) family (Gionis et al., 1999)(Charikar, 2002)(Datar et al., 2004) which use random projections to convert input data into binary hash codes. Although enjoying theoretical guarantees on sub-linear hashing/search time and the accuracy of the returned neighbors, LSH-related methods typically need long codes and a large number of hash tables to achieve good search accuracy. This may lead to considerable storage overhead and reduced search speed. Hence, in the literature, directly learning data-dependent hash functions to generate compact codes has become popular. Such hashing typically needs a small number of bits per data item and can be designed to work well with a single hash table and constant hashing time. The state-of-the-arts include unsupervised hashing (Liu et al., 2011), semi-supervised hashing (Wang et al., 2012), and supervised hashing (Liu et al., 2012).

Most of the existing hashing methods try to solve the problem of point-to-point nearest neighbor search. Namely, both queries and database items are represented as individual points in some feature space. Considering complex structures of real-world data, other forms of hashing paradigms beyond point-to-point search have also been proposed in the past, e.g., subspace-to-subspace nearest neighbor search (Basri et al., 2011). In this paper, we address a more challenging *point-to-hyperplane* search problem, where queries come as hyperplanes in $\mathbb{R}^d$, i.e., $(d-1)$-dimensional subspaces, and database items are conventional points. Then the search problem is: given a hyperplane query and a database of points, return the points which have minimal distances to the hyperplane. In the literature, not much work has been done on the point-to-hyperplane problem except (Jain et al., 2010) which demonstrated the vital importance of such a problem in making SVM-based active learning feasible on massive data pools.

Active learning (AL), also known as pool-based active learning, circumvents the high cost of blind labeling by selecting a few samples to label. At each iteration, a typical AL learner seeks the most *informative* sample from an unlabeled sample pool, so that maximal information gain is achieved after labeling the selected sample. Subsequently, the learning model is re-trained on the incrementally labeled sample set. The classical AL algorithm (Tong & Koller, 2001) used SVMs as learning models. Based on the theory of "version spaces" (Tong & Koller, 2001), it was provably shown that the best sample to select is simply the one closest to the current decision hyperplane if the assumption of symmetric version spaces holds. Unfortunately, the active selection method faces serious computational challenges when applied to gigantic databases. An exhaustive search to find the best sample is usually computationally prohibitive. Thus, fast point-to-hyperplane search is strongly desired to scale up active learning on large real-world data sets.

Recently, hyperplane hashing schemes were proposed in (Jain et al., 2010) to cope with point-to-hyperplane search. Compared with the brute-force scan through all of the database points, these schemes are significantly more efficient with theoretical guarantees of sub-linear query time and tolerable loss of accuracy for retrieved approximate nearest neighbors. Consequently, when applying hyperplane hashing to the sample selection task for SVM active learning, one can scan orders of magnitude fewer database points to deliver the next active label request, thereby making active learning scalable.

In (Jain et al., 2010), two families of randomized hash functions were proved locality-sensitive to the angle between a database point and a hyperplane query; however, long hash bits and plentiful hash tables are required to cater for the theoretical guarantees. Actually, 300 bits and 500 tables were adopted in (Jain et al., 2010) to achieve reasonable performance, which incurs a heavy burden on both computation and storage. To mitigate the above mentioned issues, this paper proposes a compact hyperplane hashing scheme which exploits only a single hash table with several tens of hash bits to tackle point-to-hyperplane search. The thrust of our hashing scheme is to design and learn *bilinear* hash functions such that nearly parallel input vectors are hashed to the same bits whereas nearly perpendicular input vectors are hashed to different bits. In fact, we first show that even without any learning, the randomized version of the proposed bilinear hashing gives higher near-neighbor collision probability than the existing methods.

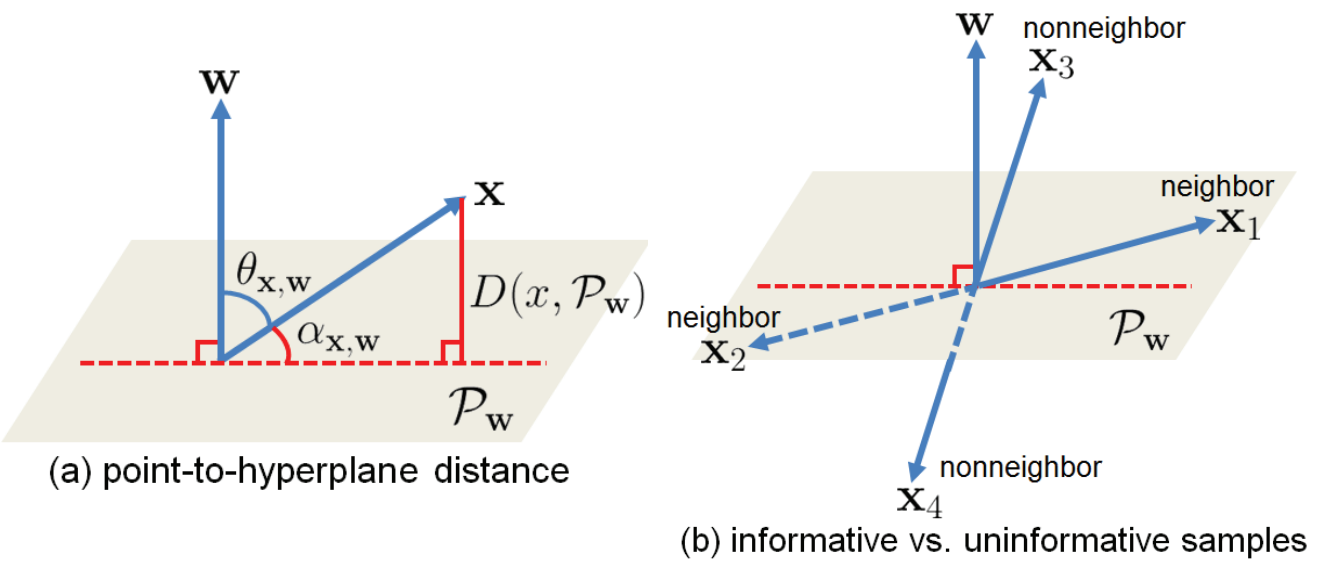

*Figure 1.* The point-to-hyperplane search problem encountered in SVM active learning. $\mathcal{P}_{\boldsymbol{w}}$ is the SVM's hyperplane decision boundary, $\boldsymbol{w}$ is the normal vector to $\mathcal{P}_{\boldsymbol{w}}$, and $\boldsymbol{x}$ is a data vector. (a) Point-to-hyperplane distance $D(\boldsymbol{x}, \mathcal{P}_{\boldsymbol{w}})$ and point-to-hyperplane angle $\alpha_{\boldsymbol{x},\boldsymbol{w}}$; (b) informative ($\boldsymbol{x}_1, \boldsymbol{x}_2$) and uninformative ($\boldsymbol{x}_3, \boldsymbol{x}_4$) samples.

Next, we cast the bilinear projections in a learning framework and show that one can do even better by using learned hash functions. Given a hyperplane query, its normal vector is used as the input and the corresponding hash code is obtained by concatenating the output bits from the learned hash functions. Then, the database points whose codes have the farthest Hamming distances to the query's code are retrieved. Critically, the retrieved points, called *near-to-hyperplane neighbors*, maintain small angles to the hyperplane following our learning principle. Experiments conducted on two large data sets up to one million corroborate that our approach enables scalable active learning with good performance. Finally, although in this paper we select SVM active learning as the testbed for hyperplane hashing, we want to highlight that the proposed compact hyperplane hashing is a general method and applicable to a large spectrum of machine learning problems such as minimal tangent distance pursuit and cutting-plane based maximum margin clustering.

## 2. Problem

First of all, let us revisit the well-known margin-based AL strategy proposed by (Tong & Koller, 2001). For the convenience of expression, we append each data vector with a 1 and use a linear kernel. Then, the SVM classifier becomes $f(\boldsymbol{x}) = \boldsymbol{w}^\top \boldsymbol{x}$ where vector $\boldsymbol{x} \in \mathbb{R}^d$ represents a data point and vector $\boldsymbol{w} \in \mathbb{R}^d$ determines a hyperplane $\mathcal{P}_{\boldsymbol{w}}$ passing through the origin. Fig. 1(a) displays the geometric relationship between $\boldsymbol{w}$ and $\mathcal{P}_{\boldsymbol{w}}$, where $\boldsymbol{w}$ is the vector normal to the hyperplane $\mathcal{P}_{\boldsymbol{w}}$. Given a hyperplane query $\mathcal{P}_{\boldsymbol{w}}$ and a database of points $\mathcal{X} = \{\boldsymbol{x}_i\}_{i=1}^n$, the active selection criterion prefers the most *informative* database point $\boldsymbol{x}^* = \arg\min_{\boldsymbol{x} \in \mathcal{X}} D(\boldsymbol{x}, \mathcal{P}_{\boldsymbol{w}})$ which has the minimum margin to the SVM's decision boundary $\mathcal{P}_{\boldsymbol{w}}$. Note that $D(\boldsymbol{x}, \mathcal{P}_{\boldsymbol{w}}) = |\boldsymbol{w}^\top \boldsymbol{x}| / \|\boldsymbol{w}\|$ is the point-to-hyperplane distance. To derive provable hyperplane hashing like (Jain et al., 2010), this paper focuses on a slightly modified "distance" $\frac{|\boldsymbol{w}^\top \boldsymbol{x}|}{\|\boldsymbol{w}\|\|\boldsymbol{x}\|}$ which is the sine of the point-to-hyperplane angle

$$\alpha_{\boldsymbol{x},\boldsymbol{w}} = \left|\theta_{\boldsymbol{x},\boldsymbol{w}} - \frac{\pi}{2}\right| = \sin^{-1}\frac{|\boldsymbol{w}^\top \boldsymbol{x}|}{\|\boldsymbol{w}\|\|\boldsymbol{x}\|}, \tag{1}$$

where $\theta_{\boldsymbol{x},\boldsymbol{w}} \in [0, \pi]$ is the angle between $\boldsymbol{x}$ and the hyperplane normal $\boldsymbol{w}$. The angle measure $\alpha_{\boldsymbol{x},\boldsymbol{w}} \in [0, \pi/2]$ between a database point and a hyperplane query can readily be reflected in hashing.

As shown in Fig. 1(b), the goal of hyperplane hashing is to hash a hyperplane query $\mathcal{P}_{\boldsymbol{w}}$ and the informative samples (e.g., $\boldsymbol{x}_1, \boldsymbol{x}_2$) with narrow $\alpha_{\boldsymbol{x},\boldsymbol{w}}$ into the same or nearby hash buckets, meanwhile avoiding to return the uninformative samples (e.g., $\boldsymbol{x}_3, \boldsymbol{x}_4$) with wide $\alpha_{\boldsymbol{x},\boldsymbol{w}}$. Because $\alpha_{\boldsymbol{x},\boldsymbol{w}} = \left|\theta_{\boldsymbol{x},\boldsymbol{w}} - \frac{\pi}{2}\right|$, the point-to-hyperplane search problem can be equivalently transformed to a specific point-to-point search problem where the query is the hyperplane normal $\boldsymbol{w}$ and the desirable nearest neighbor to the raw query $\mathcal{P}_{\boldsymbol{w}}$ is the one whose angle $\theta_{\boldsymbol{x},\boldsymbol{w}}$ from $\boldsymbol{w}$ is closest to $\pi/2$, i.e., most closely perpendicular to $\boldsymbol{w}$. This is very different from traditional point-to-point nearest neighbor search which returns the most similar point to the query point. If we regard $|\cos(\theta_{\boldsymbol{x},\boldsymbol{w}})| = \frac{|\boldsymbol{w}^\top \boldsymbol{x}|}{\|\boldsymbol{w}\|\|\boldsymbol{x}\|}$ as a similarity measure between $\boldsymbol{x}$ and $\boldsymbol{w}$, hyperplane hashing actually seeks for the most dissimilar point $\boldsymbol{x}^*$ of $|\cos(\theta_{\boldsymbol{x}^*,\boldsymbol{w}})| \to 0$ to the query point $\boldsymbol{w}$. On the contrary, the most similar point such as $\boldsymbol{w}$ or $-\boldsymbol{w}$ is surely uninformative for the active selection criterion, and must be excluded.

# 3. Randomized Hyperplane Hashing

In this section, we first briefly review the existing linear function based randomized hashing methods, then propose our bilinearly formed randomized hashing approach, and finally provide theoretic analysis for the proposed bilinear hash function.

## 3.1. Background – Linear Hash Functions

Jain *et al.* (Jain et al., 2010) devised two distinct families of randomized hash functions to attack the hyperplane hashing problem.

The first one is *Angle-Hyperplane Hash* (AH-Hash) $\mathcal{A}$, of which one example is

$$h^{\mathcal{A}}(\boldsymbol{z}) = \begin{cases} [\operatorname{sgn}(\boldsymbol{u}^\top \boldsymbol{z}), \operatorname{sgn}(\boldsymbol{v}^\top \boldsymbol{z})], & \boldsymbol{z} \text{ is a database point} \\ [\operatorname{sgn}(\boldsymbol{u}^\top \boldsymbol{z}), \operatorname{sgn}(-\boldsymbol{v}^\top \boldsymbol{z})], & \boldsymbol{z} \text{ is a hyperplane normal} \end{cases} \tag{2}$$

where $\boldsymbol{z} \in \mathbb{R}^d$ represents an input vector, and $\boldsymbol{u}$ and $\boldsymbol{v}$ are both drawn independently from a standard $d$-variate Gaussian, i.e., $\boldsymbol{u}, \boldsymbol{v} \sim \mathcal{N}(0, I_{d\times d})$. Note that $h^{\mathcal{A}}$ is a two-bit hash function which leads to the probability of collision for a hyperplane normal $\boldsymbol{w}$ and a database point $\boldsymbol{x}$:

$$\mathbf{Pr}\left[h^{\mathcal{A}}(\boldsymbol{w}) = h^{\mathcal{A}}(\boldsymbol{x})\right] = \frac{1}{4} - \frac{\alpha^2_{\boldsymbol{x},\boldsymbol{w}}}{\pi^2}. \tag{3}$$

The probability monotonically decreases as the point-to-hyperplane angle $\alpha_{\boldsymbol{x},\boldsymbol{w}}$ increases, ensuring angle-sensitive hashing.

The second is *Embedding-Hyperplane Hash* (EH-Hash) function family $\mathcal{E}$ of which one example is

$$h^{\mathcal{E}}(\boldsymbol{z}) = \begin{cases} \operatorname{sgn}\left(\mathbf{U}^\top \mathbf{V}(\boldsymbol{z}\boldsymbol{z}^\top)\right), & \boldsymbol{z} \text{ is a database point} \\ \operatorname{sgn}\left(-\mathbf{U}^\top \mathbf{V}(\boldsymbol{z}\boldsymbol{z}^\top)\right), & \boldsymbol{z} \text{ is a hyperplane normal} \end{cases} \tag{4}$$

where $\mathbf{V}(A)$ returns the vectorial concatenation of matrix $A$, and $\mathbf{U} \sim \mathcal{N}(0, I_{d^2\times d^2})$. In particular, the EH hash function $h^{\mathcal{E}}$ yields hash bits on an embedded space $\mathbb{R}^{d^2}$ resulting from vectorizing rank-one matrices $\boldsymbol{z}\boldsymbol{z}^\top$ and $-\boldsymbol{z}\boldsymbol{z}^\top$. Compared with $h^{\mathcal{A}}$, $h^{\mathcal{E}}$ gives a higher probability of collision for a hyperplane normal $\boldsymbol{w}$ and a database point $\boldsymbol{x}$:

$$\mathbf{Pr}\left[h^{\mathcal{E}}(\boldsymbol{w}) = h^{\mathcal{E}}(\boldsymbol{x})\right] = \frac{\cos^{-1}\sin^2(\alpha_{\boldsymbol{x},\boldsymbol{w}})}{\pi}, \tag{5}$$

which also bears the angle-sensitive hashing property. However, it is much more expensive to compute than AH-Hash.

It is important to note that both AH-Hash and EH-Hash are essentially linear hashing techniques. On the contrary, in this work we introduce bilinear hash functions which allow nonlinear hashing.

## 3.2. Bilinear Hash Functions

We propose a *bilinear* hash function as follows

$$h(\boldsymbol{z}) = \operatorname{sgn}(\boldsymbol{u}^\top \boldsymbol{z}\boldsymbol{z}^\top \boldsymbol{v}), \tag{6}$$

where $\boldsymbol{u}, \boldsymbol{v} \in \mathbb{R}^d$ are two projection vectors. Our motivation for devising such a bilinear form comes from the following two requirements: 1) $h$ should be invariant to the scale of $\boldsymbol{z}$, which is motivated by the fact that $\boldsymbol{z}$ and $\beta\boldsymbol{z}$ ($\beta \neq 0$) hold the same point-to-hyperplane angle; and 2) $h$ should yield different hash bits for two perpendicular input vectors. The former definitely holds due to the bilinear formulation. We show in Lemma 1 that the latter holds with a constant probability when $\boldsymbol{u}, \boldsymbol{v}$ are drawn independently from the standard normal distribution.

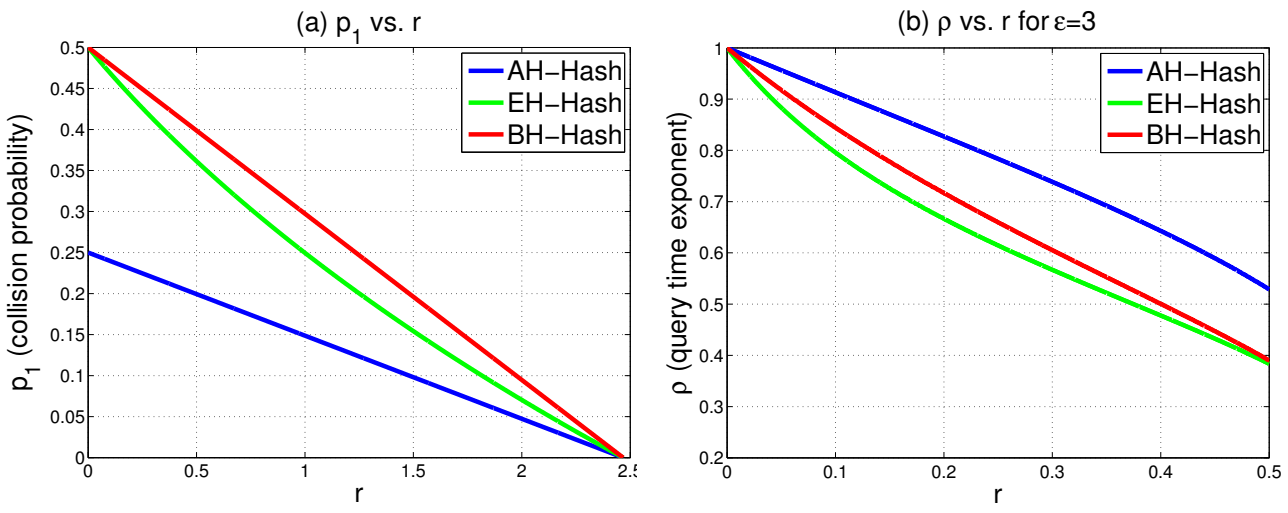

*Figure 2.* Theoretical comparison of three randomized hashing schemes. (a) $p_1$ (probability of collision) vs. $r$ (squared point-to-hyperplane angle); (b) $\rho$ (query time exponent) vs. $r$ for $\epsilon = 3$.

For the purpose of hyperplane hashing described above, the pivotal role of bilinear hash functions is to map the query point $\boldsymbol{w}$ (the hyperplane normal) and the desirable most informative point (with $\theta_{\boldsymbol{x},\boldsymbol{w}} = \pi/2$) to bitwise different hash codes, whereas map $\boldsymbol{w}$ and the undesirable most uninformative point (with $\theta_{\boldsymbol{x},\boldsymbol{w}} = 0$ or $\pi$) to identical hash codes. Therefore, hyperplane hashing works by finding the points in $\mathcal{X}$ whose codes have the largest Hamming distances to the query code of $\boldsymbol{w}$.

### 3.3. Theoretic Analysis

Based on the bilinear formulation in eq. (6), we define a novel randomized function family *Bilinear-Hyperplane Hash* (BH-Hash) as:

$$\mathcal{B} = \left\{ h^{\mathcal{B}}(\boldsymbol{z}) = \operatorname{sgn}(\boldsymbol{u}^\top \boldsymbol{z}\boldsymbol{z}^\top \boldsymbol{v}), \text{ i.i.d. } \boldsymbol{u}, \boldsymbol{v} \sim \mathcal{N}(0, I_{d\times d}) \right\}. \tag{7}$$

Here we prove several key characteristics of $\mathcal{B}$. Specially, we define $h^{\mathcal{B}}(\mathcal{P}_{\boldsymbol{w}}) = -h^{\mathcal{B}}(\boldsymbol{w})$ for an easy derivation.

**Lemma 1.** Given a hyperplane query $\mathcal{P}_{\boldsymbol{w}}$ with the normal vector $\boldsymbol{w} \in \mathbb{R}^d$ and a database point $\boldsymbol{x} \in \mathbb{R}^d$, the probability of collision for $\mathcal{P}_{\boldsymbol{w}}$ and $\boldsymbol{x}$ under $h^{\mathcal{B}}$ is

$$\mathbf{Pr}\left[h^{\mathcal{B}}(\mathcal{P}_{\boldsymbol{w}}) = h^{\mathcal{B}}(\boldsymbol{x})\right] = \frac{1}{2} - \frac{2\alpha^2_{\boldsymbol{x},\boldsymbol{w}}}{\pi^2}. \tag{8}$$

*Proof.* This probability is equal to the probability of $h^{\mathcal{B}}(\boldsymbol{w}) \neq h^{\mathcal{B}}(\boldsymbol{x})$. Because the two random projections $\boldsymbol{u}$ and $\boldsymbol{v}$ are independent,

$$\begin{aligned}&\mathbf{Pr}\left[h^{\mathcal{B}}(\boldsymbol{w}) \neq h^{\mathcal{B}}(\boldsymbol{x})\right] = \mathbf{Pr}\left[\operatorname{sgn}(\boldsymbol{u}^\top\boldsymbol{w}) = \operatorname{sgn}(\boldsymbol{u}^\top\boldsymbol{x})\right] * \\ &\mathbf{Pr}\left[\operatorname{sgn}(\boldsymbol{v}^\top\boldsymbol{w}) \neq \operatorname{sgn}(\boldsymbol{v}^\top\boldsymbol{x})\right] + \\ &\mathbf{Pr}\left[\operatorname{sgn}(\boldsymbol{u}^\top\boldsymbol{w}) \neq \operatorname{sgn}(\boldsymbol{u}^\top\boldsymbol{x})\right] \mathbf{Pr}\left[\operatorname{sgn}(\boldsymbol{v}^\top\boldsymbol{w}) = \operatorname{sgn}(\boldsymbol{v}^\top\boldsymbol{x})\right].\end{aligned}$$

By exploiting the fact $\mathbf{Pr}\left[\operatorname{sgn}(\boldsymbol{u}^\top\boldsymbol{z}) = \operatorname{sgn}(\boldsymbol{u}^\top\boldsymbol{z}')\right] = 1 - \theta_{\boldsymbol{z},\boldsymbol{z}'}/\pi$ from (Goemans & Williamson, 1995),

$$\begin{aligned}&\mathbf{Pr}\left[h^{\mathcal{B}}(\boldsymbol{w}) \neq h^{\mathcal{B}}(\boldsymbol{x})\right] \\ =&\left(1 - \frac{\theta_{\boldsymbol{x},\boldsymbol{w}}}{\pi}\right)\frac{\theta_{\boldsymbol{x},\boldsymbol{w}}}{\pi} + \frac{\theta_{\boldsymbol{x},\boldsymbol{w}}}{\pi}\left(1 - \frac{\theta_{\boldsymbol{x},\boldsymbol{w}}}{\pi}\right) \\ =&\frac{1}{2} - \frac{2(\theta_{\boldsymbol{x},\boldsymbol{w}} - \frac{\pi}{2})^2}{\pi^2} = \frac{1}{2} - \frac{2\alpha^2_{\boldsymbol{x},\boldsymbol{w}}}{\pi^2},\end{aligned}$$

which completes the proof. □

Lemma 1 shows that the probability of $h^{\mathcal{B}}(\boldsymbol{w}) \neq h^{\mathcal{B}}(\boldsymbol{x})$ is 1/2 for perpendicular $\boldsymbol{w}$ and $\boldsymbol{x}$ that hold $\theta_{\boldsymbol{x},\boldsymbol{w}} = \pi/2$ (accordingly $\alpha_{\boldsymbol{x},\boldsymbol{w}} = 0$). It is important to realize that this collision probability is twice of that from the linear AH hash function $h^{\mathcal{A}}$ described in Sec. 3.1.

**Theorem 1.** *The BH-Hash function family $\mathcal{B}$ is $\left(r, r(1+\epsilon), \frac{1}{2} - \frac{2r}{\pi^2}, \frac{1}{2} - \frac{2r(1+\epsilon)}{\pi^2}\right)$-sensitive to the distance measure $\mathcal{D}(\boldsymbol{x}, \mathcal{P}_{\boldsymbol{w}}) = \alpha^2_{\boldsymbol{x},\boldsymbol{w}}$, where $r, \epsilon > 0$.*

*Proof.* Using Lemma 1, for any $h^{\mathcal{B}} \in \mathcal{B}$, when $\mathcal{D}(\boldsymbol{x}, \mathcal{P}_{\boldsymbol{w}}) \le r$ we have

$$\begin{aligned}&\mathbf{Pr}\left[h^{\mathcal{B}}(\mathcal{P}_{\boldsymbol{w}}) = h^{\mathcal{B}}(\boldsymbol{x})\right] = \frac{1}{2} - \frac{2\mathcal{D}(\boldsymbol{x}, \mathcal{P}_{\boldsymbol{w}})}{\pi^2} \\ \ge&\frac{1}{2} - \frac{2r}{\pi^2} = p_1.\end{aligned} \tag{9}$$

Likewise, when $\mathcal{D}(\boldsymbol{x}, \mathcal{P}_{\boldsymbol{w}}) > r(1+\epsilon)$ we have

$$\mathbf{Pr}\left[h^{\mathcal{B}}(\mathcal{P}_{\boldsymbol{w}}) = h^{\mathcal{B}}(\boldsymbol{x})\right] < \frac{1}{2} - \frac{2r(1+\epsilon)}{\pi^2} = p_2. \tag{10}$$

This completes the proof. □

Note that $p_1, p_2$ ($p_1 > p_2$) depend on $0 \le r \le \pi^2/4$ and $\epsilon > 0$. We present the following theorem by adapting Theorem 1 in (Gionis et al., 1999) and Theorem 0.1 in the supplementary material of (Jain et al., 2010).

**Theorem 2.** *Suppose we have a database $\mathcal{X}$ of $n$ points. Denote the parameters $k = \log_{1/p_2} n$, $\rho = \frac{\ln p_1}{\ln p_2}$, and $c \ge 2$. Given a hyperplane query $\mathcal{P}_{\boldsymbol{w}}$, if there exists a database point $\boldsymbol{x}^*$ such that $\mathcal{D}(\boldsymbol{x}^*, \mathcal{P}_{\boldsymbol{w}}) \le r$, then the BH-Hash algorithm is able to return a database point $\hat{\boldsymbol{x}}$ such that $\mathcal{D}(\hat{\boldsymbol{x}}, \mathcal{P}_{\boldsymbol{w}}) \le r(1+\epsilon)$ with probability at least $1 - \frac{1}{c} - \frac{1}{e}$ by using $n^\rho$ hash tables of $k$ hash bits each. The query time is dominated by $O(n^\rho \log_{1/p_2} n)$ evaluations of the hash functions from $\mathcal{B}$ and $cn^\rho$ computations of the pairwise distances $\mathcal{D}$ between $\mathcal{P}_w$ and the points hashed into the same buckets.*

We defer the proof to the supplementary material due to the page limit. The query time is $O(n^\rho)$ $(0 < \rho < 1)$. For each of AH-Hash, EH-Hash and BH-Hash, we plot the collision probability $p_1$ and the query time exponent $\rho$ under $\epsilon = 3$ with varying $r$ in Fig. 2(a) and (b), respectively. At any fixed $r$, BH-Hash accomplishes the highest probability of collision, which is twice $p_1$ of AH-Hash. Though BH-Hash has slightly bigger $\rho$ than EH-Hash, it has much faster hash function computation, i.e., $\Theta(2dk)$, instead of $\Theta(d^2(k+1))$ of EH-Hash per hash table for each query or data point.

It is interesting to see that AH-Hash and our proposed BH-Hash have a tight connection in the style of hashing database points. BH-Hash actually performs the XNOR operation over the two bits that AH-Hash outputs, returning a composite single bit. As a relevant reference, the idea of applying the XOR operation over binary bits in constructing hash functions has ever been used in (Li & König, 2010). However, this is only suitable for the limited data type, discrete set, and still falls into point-to-point search.

## 4. Compact Hyperplane Hashing

Despite the higher collision probability of the proposed BH-Hash than AH-Hash and EH-Hash, it is still a randomized approach. The use of random projections in $h^{\mathcal{B}}$ has two potential issues. (i) The probability of colliding for parallel $\mathcal{P}_{\boldsymbol{w}}$ and $\boldsymbol{x}$ with $\alpha_{\boldsymbol{x},\boldsymbol{w}} = 0$ is not too high (only 1/2 according to Lemma 1). (ii) The hashing time is sub-linear $O(n^{\rho}\log_{1/p_2} n)$ in order to bound the approximation error of the retrieved neighbors, as shown in Theorem 2. AH-Hash and EH-Hash also suffer from the two issues. Even though these randomized hyperplane hashing methods maintain bounded approximation errors, they require long hash codes and plenty (even hundreds) of hash tables to cater for the accuracy guarantees. Hence, these solutions have tremendous computational and memory costs which limit the practical performance of hyperplane hashing.

To this end, we propose a *Compact Hyperplane Hashing* approach to further enhance the power of bilinear hash functions such that, instead of being random, the projections are learned from the data. Such learning yields compact yet discriminative codes which are used in a single hash table, leading to substantially reduced computational and storage needs.

We aim at learning a series of bilinear hash functions $\{h_j\}$ to yield short codes. Note that $h_j$ is different from the randomized bilinear hash function $h_j^{\mathcal{B}}$, and that we consistently define $h_j(\mathcal{P}_{\boldsymbol{w}}) = -h_j(\boldsymbol{w})$. We would like to learn $h_j$ such that smaller $\alpha_{\boldsymbol{x},\boldsymbol{w}}$ results in larger $h_j(\mathcal{P}_{\boldsymbol{w}})h_j(\boldsymbol{x})$. Thus, we make $h_j(\mathcal{P}_{\boldsymbol{w}})h_j(\boldsymbol{x})$ to monotonically decrease as $\alpha_{\boldsymbol{x},\boldsymbol{w}}$ increases. This is equivalent to the requirement that $h_j(\boldsymbol{w})h_j(\boldsymbol{x})$ monotonically increases with increasing $\sin(\alpha_{\boldsymbol{x},\boldsymbol{w}}) = |\cos(\theta_{\boldsymbol{x},\boldsymbol{w}})|$.

Suppose $k$ hash functions are learned to produce $k$-bit codes. We propose a hash function learning approach with the goal that $\sum_{j=1}^{k} h_j(\boldsymbol{w})h_j(\boldsymbol{x})/k \propto |\cos(\theta_{\boldsymbol{x},\boldsymbol{w}})|$. Further, since $\sum_{j=1}^{k} h_j(\boldsymbol{w})h_j(\boldsymbol{x})/k \in [-1,1]$ and $|\cos(\theta_{\boldsymbol{x},\boldsymbol{w}})| \in [0,1]$, we specify the learning goal as

$$\frac{1}{k}\sum_{j=1}^{k} h_j(\boldsymbol{w})h_j(\boldsymbol{x}) = 2|\cos(\theta_{\boldsymbol{x},\boldsymbol{w}})| - 1, \qquad (11)$$

which makes sense since $\theta_{\boldsymbol{x},\boldsymbol{w}} = \pi/2$, i.e., $\alpha_{\boldsymbol{x},\boldsymbol{w}} = 0$, causes $h_j(\boldsymbol{w}) \neq h_j(\boldsymbol{x})$ or $h_j(\mathcal{P}_{\boldsymbol{w}}) = h_j(\boldsymbol{x})$ for any $j \in [1:k]$. As such, the proposed learning method achieves explicit collision for parallel $\mathcal{P}_{\boldsymbol{w}}$ and $\boldsymbol{x}$.

Enforcing eq. (11) tends to make $h_j$ to yield identical hash codes for nearly parallel inputs whereas bitwise different hash codes for nearly perpendicular inputs. At the query time, given a hyperplane query, we first extract its $k$-bit hash code using the $k$ learned hash functions applied to the hyperplane normal vector. Then, the database points whose codes have the largest Hamming distances to the query's code are returned. Thus, the returned points, called *near-to-hyperplane neighbors*, maintain small angles to the hyperplane because such points and the hyperplane normal are nearly perpendicular. In our learning setting, $k$ is typically very short, no more than 30, so we can retrieve the desirable near-to-hyperplane neighbors via constant time hashing over a single hash table.

Now we desribe how we learn $k$ pairs of projections $(\boldsymbol{u}_j, \boldsymbol{v}_j)_{j=1}^{k}$ so as to construct $k$ bilinear hash functions $\{h_j(\boldsymbol{z}) = \mathrm{sgn}(\boldsymbol{u}_j^\top \boldsymbol{z}\boldsymbol{z}^\top \boldsymbol{v}_j)\}_{j=1}^{k}$. Since the hyperplane normal vectors come up only during the query time, we cannot access $\boldsymbol{w}$ during the training stage. Instead, we sample a few database points for learning projections. Without the loss of generality, we assume that the first $m$ ($k < m \ll n$) samples saved in the matrix $X_m = [\boldsymbol{x}_1, \cdots, \boldsymbol{x}_m]$ are used for learning. To capture the pairwise relationships among them, we define a matrix $S \in \mathbb{R}^{m\times m}$ as

$$S_{ii'} = \begin{cases} 1, & |\cos(\theta_{\boldsymbol{x}_i,\boldsymbol{x}_{i'}})| \geq t_1 \\ -1, & |\cos(\theta_{\boldsymbol{x}_i,\boldsymbol{x}_{i'}})| \leq t_2 \\ 2|\cos(\theta_{\boldsymbol{x}_i,\boldsymbol{x}_{i'}})| - 1, & \text{otherwise} \end{cases} \qquad (12)$$

where $0 < t_2 < t_1 < 1$ are two thresholds. For any sample $\boldsymbol{x}$, its $k$-bit hash code is written as $H(\boldsymbol{x}) = [h_1(\boldsymbol{x}), \cdots, h_k(\boldsymbol{x})]$, and $\sum_{j=1}^{k} h_j(\boldsymbol{x}_i)h_j(\boldsymbol{x}_{i'}) = H(\boldsymbol{x}_i)H^\top(\boldsymbol{x}_{i'})$. By taking advantage of the learning goal given in eq. (11), we formulate a least-squares style objective function $\mathcal{Q}$ to learn $X_m$'s binary codes as $\mathcal{Q} = \left\|\frac{1}{k}BB^\top - S\right\|_{\mathrm{F}}^2$, where $B = [H^\top(\boldsymbol{x}_1), \cdots, H^\top(\boldsymbol{x}_m)]^\top$ represents the code matrix

of $X_m$ and $\|.\|_{\mathrm{F}}$ denotes the Frobenius norm. The thresholds $t_1, t_2$ used in eq. (12) have an important role. When two inputs are prone to being parallel so that $|\cos(\theta_{\boldsymbol{x}_i,\boldsymbol{x}_{i'}})|$ is large enough ($\geq t_1$), minimizing $\mathcal{Q}$ drives each bit of their codes to collide, i.e., $H(\boldsymbol{x}_i)H^\top(\boldsymbol{x}_{i'})/k = 1$; when two inputs tend to be perpendicular so that $|\cos(\theta_{\boldsymbol{x}_i,\boldsymbol{x}_{i'}})|$ is small enough ($\leq t_2$), minimizing $\mathcal{Q}$ tries to make their codes bit-by-bit different, i.e., $H(\boldsymbol{x}_i)H^\top(\boldsymbol{x}_{i'})/k = -1$.

With simple algebra, one can rewrite $\mathcal{Q}$ as

$$\min_{(\boldsymbol{u}_j,\boldsymbol{v}_j)_{j=1}^k} \left\| \sum_{j=1}^k \boldsymbol{b}_j\boldsymbol{b}_j^\top - kS \right\|_{\mathrm{F}}^2$$
$$\text{s.t.} \quad \boldsymbol{b}_j = \begin{bmatrix} \mathrm{sgn}\left(\boldsymbol{u}_j^\top \boldsymbol{x}_1 \boldsymbol{x}_1^\top \boldsymbol{v}_j\right) \\ \cdots\cdots \\ \mathrm{sgn}\left(\boldsymbol{u}_j^\top \boldsymbol{x}_m \boldsymbol{x}_m^\top \boldsymbol{v}_j\right) \end{bmatrix}. \tag{13}$$

Every bit vector $\boldsymbol{b}_j \in \{1,-1\}^m$ in $B = [\boldsymbol{b}_1, \cdots, \boldsymbol{b}_k]$ determines one hash function $h_j$ parameterized by one projection pair $(\boldsymbol{u}_j, \boldsymbol{v}_j)$. Note that $\boldsymbol{b}_j$'s are separable in the summation, which inspires a greedy idea for solving $\boldsymbol{b}_j$'s sequentially. At a time, it only involves solving one bit vector $\boldsymbol{b}_j(\boldsymbol{u}_j, \boldsymbol{v}_j)$ given the previously solved vectors $\boldsymbol{b}_1^*, \cdots, \boldsymbol{b}_{j-1}^*$. Let us define a residue matrix $R_{j-1} = kS - \sum_{j'=1}^{j-1} \boldsymbol{b}_{j'}\boldsymbol{b}_{j'}^\top$ with $R_0 = kS$. Then, $\boldsymbol{b}_j$ can be pursued by minimizing the following cost

$$\begin{aligned}\left\|\boldsymbol{b}_j\boldsymbol{b}_j^\top - R_{j-1}\right\|_{\mathrm{F}}^2 &= \left(\boldsymbol{b}_j^\top\boldsymbol{b}_j\right)^2 - 2\boldsymbol{b}_j^\top R_{j-1}\boldsymbol{b}_j + \mathrm{tr}(R_{j-1}^2) \\ &= -2\boldsymbol{b}_j^\top R_{j-1}\boldsymbol{b}_j + m^2 + \mathrm{tr}(R_{j-1}^2) \\ &= -2\boldsymbol{b}_j^\top R_{j-1}\boldsymbol{b}_j + const. \end{aligned} \tag{14}$$

Discarding the constant term, the final cost is given as

$$g(\boldsymbol{u}_j, \boldsymbol{v}_j) = -\boldsymbol{b}_j^\top R_{j-1}\boldsymbol{b}_j. \tag{15}$$

Note that $g(\boldsymbol{u}_j, \boldsymbol{v}_j)$ is lower-bounded as eq. (14) is always nonnegative. However, minimizing $g$ is not easy because it is neither convex nor smooth. Below we propose an approximate optimization algorithm.

Since the hardness of minimizing $g$ lies in the sign function, we replace sgn() in $\boldsymbol{b}_j$ with the sigmoid-shaped function $\varphi(x) = 2/(1+\exp(-x)) - 1$ which is sufficiently smooth and well approximates $\mathrm{sgn}(x)$ when $|x| > 6$. Subsequently, we propose to optimize a smooth surrogate $\tilde{g}$ of $g$ defined by

$$\tilde{g}(\boldsymbol{u}_j, \boldsymbol{v}_j) = -\tilde{\boldsymbol{b}}_j^\top R_{j-1}\tilde{\boldsymbol{b}}_j, \tag{16}$$

where the vector

$$\tilde{\boldsymbol{b}}_j = \begin{bmatrix} \varphi\left(\boldsymbol{u}_j^\top \boldsymbol{x}_1 \boldsymbol{x}_1^\top \boldsymbol{v}_j\right) \\ \cdots\cdots \\ \varphi\left(\boldsymbol{u}_j^\top \boldsymbol{x}_m \boldsymbol{x}_m^\top \boldsymbol{v}_j\right) \end{bmatrix}. \tag{17}$$

We derive the gradient of $\tilde{g}$ with respect to $[\boldsymbol{u}_j^\top, \boldsymbol{v}_j^\top]^\top$

$$\nabla\tilde{g} = -\begin{bmatrix} X_m\Sigma_j X_m^\top \boldsymbol{v}_j \\ X_m\Sigma_j X_m^\top \boldsymbol{u}_j \end{bmatrix}, \tag{18}$$

where $\Sigma_j \in \mathbb{R}^{m\times m}$ is a diagonal matrix whose diagonal elements come from the $m$-dimensional vector $(R_{j-1}\tilde{\boldsymbol{b}}_j) \odot (\mathbf{1} - \tilde{\boldsymbol{b}}_j \odot \tilde{\boldsymbol{b}}_j)$. Here the symbol $\odot$ represents the Hadamard product (i.e., elementwise product), and $\mathbf{1}$ denotes a constant vector with $m$ 1 entries. Since the original cost $g$ in eq. (15) is lower-bounded, its smooth surrogate $\tilde{g}$ in eq. (16) is lower-bounded as well. We are thus able to minimize $\tilde{g}$ using the regular gradient descent technique. Note that the smooth surrogate $\tilde{g}$ is still nonconvex, so it is unrealistic to look for a global minima of $\tilde{g}$. For fast convergence, we adopt a pair of random projections $(\boldsymbol{u}_j^0, \boldsymbol{v}_j^0)$, which were used in $h_j^{\mathcal{B}}$, as a warm start and apply Nesterov's gradient method (Nesterov, 2003) to accelerate the gradient decent procedure. In most cases we attain a locally optimal $(\boldsymbol{u}_j^*, \boldsymbol{v}_j^*)$ at which $\tilde{g}(\boldsymbol{u}_j^*, \boldsymbol{v}_j^*)$ is very close to its lower bound.

The final optimized bilinear hash functions are given as $\left\{h_j(\boldsymbol{z}) = \mathrm{sgn}\left((\boldsymbol{u}_j^*)^\top \boldsymbol{z}\boldsymbol{z}^\top \boldsymbol{v}_j^*\right)\right\}_{j=1}^k$. Although, unlike the randomized hashing, it is not easy to prove their theoretical properties such as the collision probability, they result in a more accurate point-to-hyperplane search than the randomized functions $\left\{h_j^{\mathcal{B}}\right\}$, as demonstrated by the subsequent experiments.

With the learned hash functions $H = [h_1, \cdots, h_k]$ in hand, we can implement the proposed compact hyperplane hashing by simply treating a -1 bit as a 0 bit. In the preprocessing stage, each database point $\boldsymbol{x}$ is converted into a $k$-bit hash code $H(\boldsymbol{x})$ and stored in a single hash table with $k$-bit hash keys as entries. To perform search at the query time, given a hyperplane normal $\boldsymbol{w}$, we 1) extract its hash key $H(\boldsymbol{w})$ and perform the bitwise NOT operation to get the key $\overline{H(\boldsymbol{w})}$; 2) look up $\overline{H(\boldsymbol{w})}$ in the hash table for the nearest entries up to a small Hamming distance, obtaining a short list $\mathcal{L}$ whose points are retrieved from the found hash buckets; 3) scan the list $\mathcal{L}$ and then return the point $x^* = \arg\min_{\boldsymbol{x}\in\mathcal{L}} |\boldsymbol{w}^\top\boldsymbol{x}|/\|\boldsymbol{w}\|$. In fact, searching within a small Hamming ball centered at the flipped code $\overline{H(\boldsymbol{w})}$ is equivalent to searching the codes that have largest possible Hamming distances to the code $H(\boldsymbol{w})$ in the Hamming space.

## 5. Experiments

### 5.1. Datasets

We conduct experiments on two publicly available datasets including the **20 Newsgroups** textual corpus and the 1.06 million subset, called **Tiny-1M**, of the 80 million tiny image collection[1]. The first dataset is the version 2[2] of **20 Newsgroups**. It is comprised of 18,846 documents from 20 newsgroup categories. Each document is represented by a 26,214-dimensional *tf-idf* feature vector that is $\ell_2$ normalized. The **Tiny-1M** dataset is a union of **CIFAR-10**[3] and one million tiny images sampled from the entire 80M tiny image set. **CIFAR-10** is a labeled subset of the 80M tiny image set, consisting of a total of 60,000 color images from ten object categories each of which has 6000 samples. The other 1M images do not have annotated labels. In our experiments, we treat them as the "other" class besides the ten classes appearing in **CIFAR-10**, since they were sampled as the farthest 1M images to the mean image of **CIFAR-10**. Each image in **Tiny-1M** is represented by a 384-dimensional GIST (Oliva & Torralba, 2001) feature vector.

For each dataset, we train a linear SVM in the one-versus-all setting with an initially labeled set which contains randomly selected labeled samples from all classes, and then run active sample selection for 300 iterations. The initially labeled set for **20 Newsgroups** includes 5 samples per class, while for **Tiny-1M** includes 50 samples per class. For both datasets, we try 5 random initializations. After each sample selection is made, we add it to the labeled set and re-train the SVM. We use LIBLINEAR[4] for running linear SVMs. All our experiments were run on a workstation with a 2.53 GHz Intel Xeon CPU and 48GB RAM.

### 5.2. Evaluations and Results

We carry out SVM active learning using the minimum-margin based sample selection criterion for which we apply hyperplane hashing techniques to expedite the selection procedure. To validate the actual performance of the discussed hyperplane hashing methods, we compare them with two baselines: *random selection* where the next label request is randomly made, and *exhaustive selection* where the margin criterion is evaluated for all currently unlabeled samples. We compare four hashing methods including two randomized linear hashing schemes AH-Hash and EH-Hash (Jain et al., 2010), the proposed randomized bilinear hashing scheme BH-Hash, and the proposed learning-based bilinear hashing scheme that we call LBH-Hash. Notice that we use the same random projections for AH-Hash, BH-Hash, and the initialization of LBH-Hash to shed light on the effect of bilinear hashing (XNOR two bits). We also follow the dimension-sampling trick in (Jain et al., 2010) to accelerate EH-Hash's computation. In order to train our proposed LBH-Hash, we randomly sample 500 and 5000 database points from **20 Newsgroups** and **Tiny-1M**, respectively. The two thresholds $t_1, t_2$ used for implementing explicit collision are acquired according to the following rule: compute the absolute cosine matrix $C$ between the $m$ sampled points $\{\boldsymbol{x}_i\}_{i=1}^m$ and all data, average the top 5% values among $C_{i.}$ across $\boldsymbol{x}_i$'s as $t_1$, and average the bottom 5% values as $t_2$.

So as to make the hashing methods work under a compact hashing mode for fair comparison, we employ a single hash table with short code length. Concretely, we use 16 hash bits for EH-Hash, BH-Hash, and LBH-Hash, and 32 bits for AH-Hash because of its dual-bit hashing spirit on **20 Newsgroups**. When applying each hashing method in an AL iteration, we perform a hash lookup within Hamming radius 3 in the corresponding hash table and then scan the points in the found hash buckets, resulting in the neighbor near to the current SVM's decision hyperplane. Likewise, we use 20 bits for EH-Hash, BH-Hash, and LBH-Hash, and 40 bits for AH-Hash on **Tiny-1M**; the Hamming radius for search is set to 4. It is possible that a method finds all empty hash buckets in the Hamming ball. In that case, we apply random selection as a supplement.

We evaluate the performance of four hashing methods in terms of: 1) the average precision (AP) which is computed by ranking the current unlabeled sample set with the current SVM classifier at each AL iteration; 2) the minimum margin (the smallest point-to-hyperplane distance $|\boldsymbol{w}^\top \boldsymbol{x}|/\|\boldsymbol{w}\|$) of the neighbor returned by hyperplane hashing at each AL iteration; 3) the number of queries among a total of 300 for every class that receive nonempty hash lookups. The former two results are averaged over all classes and 5 runs, and the latter is averaged over 5 runs. We report such results in Fig. 3 and Fig. 4, which clearly show that 1) LBH-Hash achieves the highest mean AP (MAP) among all compared hashing methods, and even outperforms exhaustive selection at some AL iterations; 2) LBH-Hash accomplishes the minimum margin closest to that by exhaustive selection; 3) LBH-Hash enjoys almost all nonempty hash lookups (AH-Hash gets almost all empty lookups). The superior performance of LBH-Hash corroborates that the proposed bilinear hash function and the associated learning technique are successful in utilizing the underlying data information to yield compact yet discriminative codes.

Finally, we report the computational efficiency in Tables 1-3 of the supplementary material, which indicate

[1] http://horatio.cs.nyu.edu/mit/tiny/data/index.html
[2] http://www.zjucadcg.cn/dengcai/Data/TextData.html
[3] http://www.cs.toronto.edu/~kriz/cifar.html
[4] http://www.csie.ntu.edu.tw/~cjlin/liblinear/

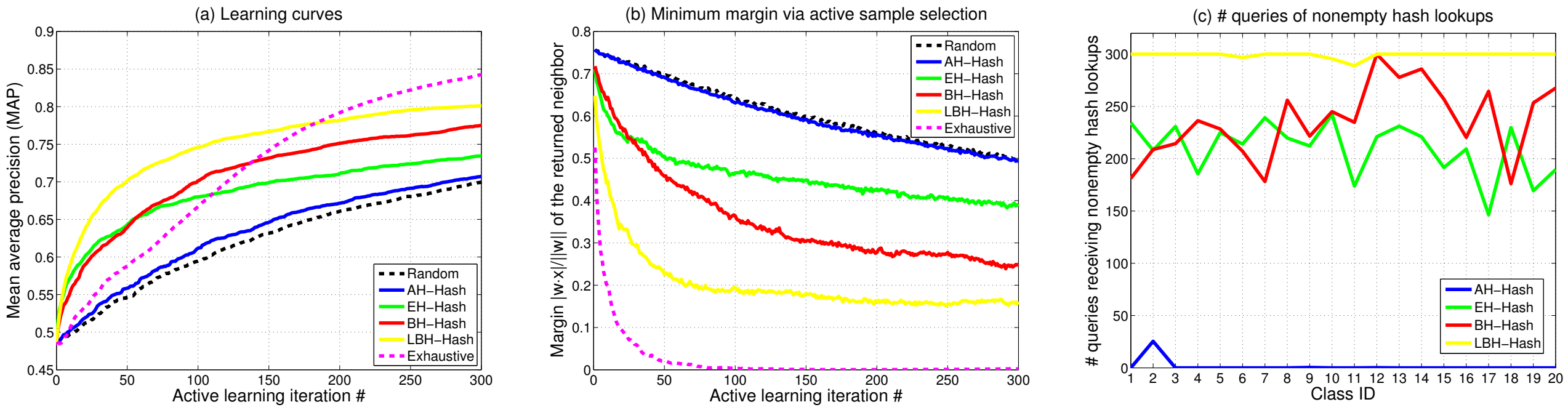

*Figure 3.* Results on **20 Newsgroups**. (a) Learning curves of MAP, (b) minimum-margin curves of active sample selection, and (c) the number of queries ($\leq 300$) receiving nonempty hash lookups across 20 classes.

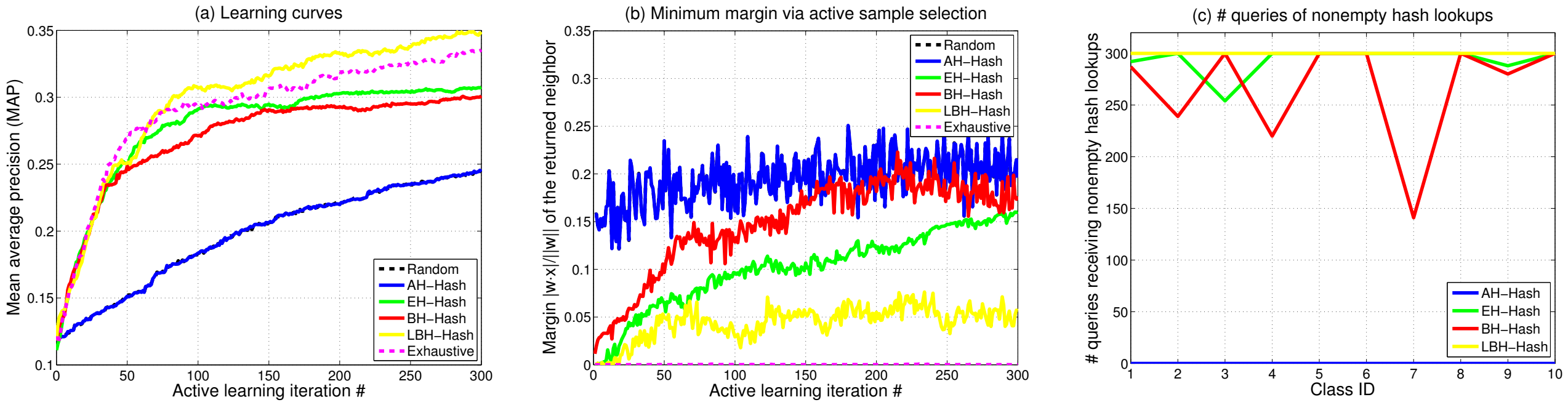

*Figure 4.* Results on **Tiny-1M**. (a) Learning curves of MAP, (b) minimum-margin curves of active sample selection, and (c) the number of queries ($\leq 300$) receiving nonempty hash lookups across 10 classes.

that LBH-Hash takes comparable preprocessing time as EH-Hash and achieves fast search speed.

## 6. Conclusions

We have addressed hyperplane hashing by proposing a specialized bilinear hash function which allows efficient search of points near a hyperplane query. Even when using random projections, the proposed hash function enjoys higher probability of collision than the existing randomized methods. By learning the projections further, we achieve compact yet discriminative codes that permit substantial savings in both storage and time needed during search. Large-scale active learning experiments on two datasets have demonstrated the superior performance of our compact hyperplane hashing approach.

**Acknowledgement:** This work is supported in part by a Facebook fellowship to the first author.